\pdfoutput=1

\documentclass[11pt]{article}

\usepackage[]{acl}

\usepackage{times}
\usepackage{latexsym}

\usepackage[T1]{fontenc}

\usepackage[utf8]{inputenc}

\usepackage{microtype}

\usepackage{bm}
\usepackage{amsmath}
\usepackage{url}
\usepackage{graphicx}
\usepackage{float} 
\usepackage{subfigure}
\usepackage{multirow}
\usepackage{makecell}
\usepackage{xpinyin}
\usepackage{enumerate}

\usepackage{xcolor}
\usepackage{soul}
\newcolumntype{L}[1]{>{\raggedright\let\newline\\\arraybackslash\hspace{0pt}}m{#1}}
\setul{2pt}{2pt}
\usepackage{booktabs}
\usepackage{amsmath,amsfonts}
\usepackage[utf8]{inputenc}
\usepackage{cleveref}

\crefname{section}{§}{§§}
\Crefname{section}{§}{§§}
\usepackage{enumitem}
\setenumerate[1]{itemsep=0pt,partopsep=0pt,parsep=\parskip,topsep=5pt}
\setitemize[1]{itemsep=0pt,partopsep=0pt,parsep=\parskip,topsep=5pt}
\setdescription{itemsep=0pt,partopsep=0pt,parsep=\parskip,topsep=5pt}
\usepackage{arydshln}

\usepackage[linesnumbered,ruled,vlined]{algorithm2e}

\title{A Variational Hierarchical Model for Neural Cross-Lingual Summarization}

\author{
  Yunlong Liang\textsuperscript{1}\thanks{ \ \ Work was done when Liang and Zhou were interning at Pattern Recognition Center, WeChat AI, Tencent Inc, China.}  , 
  Fandong Meng\textsuperscript{2}, 
  Chulun Zhou\textsuperscript{2,3},
  \textbf{Jinan Xu}\textsuperscript{1}\thanks{ \ \ Jinan Xu is the corresponding author.}  , \\
  \textbf{Yufeng Chen}\textsuperscript{1}, 
  \textbf{Jinsong Su}\textsuperscript{3}
  and \textbf{Jie Zhou}\textsuperscript{2}\\
  \textsuperscript{1}Beijing Key Lab of Traffic Data Analysis and Mining, Beijing Jiaotong University, China \\
  \textsuperscript{2}Pattern Recognition Center, WeChat AI, Tencent Inc, China \\
  \textsuperscript{3}School of Informatics, Xiamen University \\
  \texttt{\{yunlongliang,jaxu,chenyf\}@bjtu.edu.cn} \ \ \ \  \texttt{clzhou@stu.xmu.edu.cn}\\
  \texttt{jssu@xmu.edu.cn} \ \ \ \  \texttt{\{fandongmeng,withtomzhou\}@tencent.com} \\
}

\begin{document}
\maketitle
\begin{abstract}
The goal of the cross-lingual summarization (CLS) is to convert a document in one language (\emph{e.g.}, English) to a summary in another one (\emph{e.g.}, Chinese). Essentially, the CLS task is the combination of machine translation (MT) and monolingual summarization (MS), and thus there exists the hierarchical relationship between MT\&MS and CLS. Existing studies on CLS mainly focus on utilizing pipeline methods or jointly training an end-to-end model through an auxiliary MT or MS objective. However, it is very challenging for the model to directly conduct CLS as it requires both the abilities to translate and summarize. To address this issue, we propose a hierarchical model for the CLS task, based on the conditional variational auto-encoder. The hierarchical model contains two kinds of latent variables at the local and global levels, respectively. At the local level, there are two latent variables, one for translation and the other for summarization. As for the global level, there is another latent variable for cross-lingual summarization conditioned on the two local-level variables. Experiments on two language directions (English$\Leftrightarrow$Chinese) verify the effectiveness and superiority of the proposed approach. In addition, we show that our model is able to generate better cross-lingual summaries than comparison models in the few-shot setting. 
\end{abstract}

\section{Introduction}
The cross-lingual summarization (CLS) aims to summarize a document in source language (\emph{e.g.}, English) into a different language (\emph{e.g.}, Chinese), which can be seen as a combination of machine translation (MT) and monolingual summarization (MS) to some extent~\cite{orasan-chiorean-2008-evaluation,zhu-etal-2019-ncls}. The CLS can help people effectively master the core points of an article in a foreign language. Under the background of globalization, it becomes more important and is now coming into widespread use in real life.

Many researches have been devoted to dealing with this task. To our knowledge, they mainly fall into two categories, \emph{i.e.}, pipeline and end-to-end learning methods. (\emph{\romannumeral1}) The first category is pipeline-based, adopting either translation-summarization~\cite{10.1145/979872.979877,ouyang-etal-2019-robust} or summarization-translation~\cite{wan-etal-2010-cross,orasan-chiorean-2008-evaluation} paradigm. Although being intuitive and straightforward, they generally suffer from error propagation problem. 
(\emph{\romannumeral2}) The second category aims to train an end-to-end model for CLS~\cite{zhu-etal-2019-ncls,zhu-etal-2020-attend}. For instance, ~\citet{zhu-etal-2020-attend} focus on using a pre-constructed probabilistic bilingual lexicon to improve the CLS model. Furthermore, some researches resort to multi-task learning~\cite{takase2020multitask,bai-etal-2021-cross,zhu-etal-2019-ncls,cao-etal-2020-jointly,Cao_Wan_Yao_Yu_2020}. \citet{zhu-etal-2019-ncls} separately introduce MT and MS to improve CLS. \citet{cao-etal-2020-jointly,Cao_Wan_Yao_Yu_2020} design several additional training objectives (\emph{e.g.}, MS, back-translation, and reconstruction) to enhance the CLS model. And \citet{xu-etal-2020-mixed} utilize a mixed-lingual pre-training method with several auxiliary tasks for CLS.

As pointed out by~\citet{cao-etal-2020-jointly}, it is challenging for the model to directly conduct CLS as it requires both the abilities to translate and summarize. Although some methods have used the related tasks (\emph{e.g.}, MT and MS) to help the CLS, the hierarchical relationship between MT\&MS and CLS are not well modeled, which can explicitly enhance the CLS task. Apparently, how to effectively model the hierarchical relationship to exploit MT and MS is one of the core issues, especially when the CLS data are limited.\footnote{Generally, it is difficult to acquire the CLS dataset~\cite{zhu-etal-2020-attend,liu2018zero,duan-etal-2019-zero}.} In many other related NLP tasks~\cite{park-etal-2018-hierarchical,AAAI1714567,shen-etal-2019-modeling,shen-etal-2021-gtm}, the Conditional Variational Auto-Encoder (CVAE)~\cite{NIPS2015_8d55a249} has shown its superiority in learning hierarchical structure with hierarchical latent variables, which is often leveraged to capture the semantic connection between the utterance and the corresponding context of conversations. Inspired by these work, we attempt to adapt CVAE to model the hierarchical relationship between MT\&MS and CLS. 

Therefore, we propose a {V}ariational {H}ierarchical {M}odel to exploit translation and summarization simultaneously, named VHM, for CLS task in an end-to-end framework. VHM employs hierarchical latent variables based on CVAE to learn the hierarchical relationship between MT\&MS and CLS. Specifically, the VHM contains two kinds of latent variables at the local and global levels, respectively. Firstly, we introduce two local variables for translation and summarization, respectively. The two local variables are constrained to reconstruct the translation and source-language summary. Then, we use the global variable to explicitly exploit the two local variables for better CLS, which is constrained to reconstruct the target-language summary.  
This makes sure the global variable captures its relationship with the two local variables without any loss, preventing error propagation. 
For inference, we use the local and global variables to assist the cross-lingual summarization process.

We validate our proposed training framework on the datasets of different language pairs~\cite{zhu-etal-2019-ncls}: Zh2EnSum (Chinese$\Rightarrow$English) and En2ZhSum (English$\Rightarrow$Chinese). Experiments show that our model achieves consistent improvements on two language directions in terms of both automatic metrics and human evaluation, demonstrating its effectiveness and generalizability. Few-shot evaluation further suggests that the local and global variables enable our model to generate a satisfactory cross-lingual summaries compared to existing related methods.  

Our main contributions are as follows\footnote{The code is publicly available at: \url{https://github.com/XL2248/VHM}}:
\begin{itemize}

\item We are the first that builds a variational hierarchical model via conditional variational auto-encoders that introduce a global variable to combine the local ones for translation and summarization at the same time for CLS.

\item Our model gains consistent and significant performance and remarkably outperforms the most previous state-of-the-art methods after using mBART~\cite{liu-etal-2020-multilingual-denoising}.

\item Under the few-shot setting, our model still achieves better performance than existing approaches. Particularly, the fewer the data are, the greater the improvement we gain.

\end{itemize}

\section{Background}
\paragraph{Machine Translation (MT).}
Given an input sequence in the source language $X_{mt}$$=$$\{x_i\}_{i=1}^{|X_{mt}|}$, the goal of the neural MT model is to produce its translation in the target language $Y_{mt}$$=$$\{y_i\}_{i=1}^{|Y_{mt}|}$. The conditional distribution of the model is:
\begin{equation}\nonumber
\setlength{\abovedisplayskip}{5pt}
\setlength{\belowdisplayskip}{5pt}
\label{eq:nmt}
    p_{\theta}(Y_{mt}|X_{mt}) = \prod_{t=1}^{|Y_{mt}|}p_{\theta}(y_t|X_{mt}, y_{1:t-1}),
\end{equation}
where $\theta$ are model parameters and $y_{1:t-1}$ is the partial translation. 

\paragraph{Monolingual Summarization (MS).}
Given an input article in the source language $X_{ms}^{src}$$=$$\{x_i^{src}\}_{i=1}^{|X_{ms}^{src}|}$ and the corresponding summarization in the same language $X_{ms}^{tgt}$$=$$\{x_i^{tgt}\}_{i=1}^{|X_{ms}^{tgt}|}$, the monolingual summarization is formalized as:
\begin{equation}\nonumber
\setlength{\abovedisplayskip}{5pt}
\setlength{\belowdisplayskip}{5pt}
\label{eq:ms}
    p_{\theta}(X_{ms}^{tgt}|X_{ms}^{src}) = \prod_{t=1}^{|X_{ms}^{tgt}|}p_{\theta}(x_t^{tgt}|X_{ms}^{src}, x_{1:t-1}^{tgt}).
\end{equation}

\paragraph{Cross-Lingual Summarization (CLS). }
In CLS, we aim to learn a model that can generate a summary in the target language $Y_{cls}$$=$$\{y_i\}_{i=1}^{|Y_{cls}|}$ for a given article in the source language $X_{cls}$$=$$\{x_i\}_{i=1}^{|X_{cls}|}$. Formally, it is as follows: 
\begin{equation}\nonumber
\setlength{\abovedisplayskip}{5pt}
\setlength{\belowdisplayskip}{5pt}
\label{eq:cls}
    p_{\theta}(Y_{cls}|X_{cls}) = \prod_{t=1}^{|Y_{cls}|}p_{\theta}(y_t|X_{cls}, y_{1:t-1}).
\end{equation}

\noindent \textbf{Conditional Variational Auto-Encoder (CVAE).}
The CVAE~\cite{NIPS2015_8d55a249} consists of one prior network and one recognition (posterior) network, where the latter takes charge of guiding the learning of prior network via Kullback–Leibler (KL) divergence~\cite{kingma2013auto}. For example, the variational neural MT model~\cite{zhang-etal-2016-variational,Su_Wu_Xiong_Lu_Han_Zhang_2018,mccarthy-etal-2020-addressing,8246560}, which introduces a random latent variable $\mathbf{z}$ into the neural MT conditional distribution:
\begin{equation}
\setlength{\abovedisplayskip}{5pt}
\setlength{\belowdisplayskip}{5pt}
\label{eq:vnmt}
    p_{\theta}(Y_{mt}|X_{mt}) = \int_\mathbf{z} p_{\theta}(Y_{mt}|X_{mt}, \mathbf{z}) \cdot p_{\theta}(\mathbf{z}|X_{mt}) d\mathbf{z}.
\end{equation}
Given a source sentence $X$, a latent variable $\mathbf{z}$ is firstly sampled by the prior network from the encoder, and then the target sentence is generated by the decoder: $Y_{mt} \sim p_{\theta}(Y_{mt}|X_{mt}, \mathbf{z})$, where $\mathbf{z} \sim p_{\theta}(\mathbf{z}|X_{mt})$. 

As it is hard to marginalize \autoref{eq:vnmt}, the CVAE training objective is a variational lower bound of the conditional log-likelihood:
\begin{equation}
\resizebox{.98\hsize}{!}{$
\begin{split}   
\label{eq:elbo}
    \mathcal{L}(\theta,\phi;X_{mt},Y_{mt}) &= -\mathrm{KL}(q_\phi (\mathbf{z}^{\prime}|X_{mt},Y_{mt}) \| p_\theta (\mathbf{z}|X_{mt})) \nonumber \\
                       &+ \mathbb{E}_{q_\phi (\mathbf{z}^{\prime}|X_{mt},Y_{mt})} [\log p_\theta(Y_{mt}|\mathbf{z}, X_{mt})] \\
                       & \leq \log p (Y_{mt}|X_{mt}), \nonumber
\end{split}
$}
\end{equation}
where $\phi$ are parameters of the CVAE.

\section{Methodology}
\autoref{fig:model} demonstrates an overview of our model, consisting of four components: \emph{encoder}, \emph{variational hierarchical modules}, \emph{decoder}, \emph{training and inference}. Specifically, we aim to explicitly exploit the MT and MS for CLS simultaneously. Therefore, we firstly use the \emph{encoder} (\autoref{sec:enc}) to prepare the representation for the \emph{variational hierarchical module} (\autoref{sec:rsc}), which aims to learn the two local variables for the global variable in CLS. Then, we introduce the global variable into the \emph{decoder} (\autoref{sec:dec}). Finally, we elaborate the process of our \emph{training and inference} (\autoref{sec:to}).

\subsection{Encoder}
\label{sec:enc}
Our model is based on transformer~\cite{vaswani2017attention} framework. As shown in~\autoref{fig:model}, the encoder takes six types of inputs, \{$X_{mt}$, $X_{ms}^{src}$, $X_{cls}$, $Y_{mt}$, $X_{ms}^{tgt}$,  $Y_{cls}$\}, among which  $Y_{mt}$, $X_{ms}^{tgt}$, and $Y_{cls}$ are only for training recognition networks. Taking $X_{mt}$ for example, the encoder maps the input $X_{mt}$ into a sequence of continuous representations whose size varies with respect to the source sequence length. Specifically, the encoder consists of $N_e$ stacked layers and each layer includes two sub-layers:\footnote{The layer normalization is omitted for simplicity and you may
refer to~\cite{vaswani2017attention} for more details.} a multi-head self-attention ($\mathrm{SelfAtt}$) sub-layer and a position-wise feed-forward network ($\mathrm{FFN}$) sub-layer:
\begin{equation}
\setlength{\abovedisplayskip}{5pt}
\setlength{\belowdisplayskip}{5pt}
\begin{split}
    \mathbf{s}^\ell_e &= \mathrm{SelfAtt}(\mathbf{h}^{\ell-1}_e) + \mathbf{h}^{\ell-1}_e, \nonumber\\
    \mathbf{h}^\ell_e &= \mathrm{FFN}(\mathbf{s}^\ell_e) + \mathbf{s}^\ell_e, \nonumber
\end{split}
\end{equation}
where $\mathbf{h}^{\ell}_e$ denotes the state of the $\ell$-th encoder layer and $\mathbf{h}^{0}_e$ denotes the initialized embedding. 

Through the encoder, we prepare the representations of \{$X_{mt}$, $X_{ms}^{src}$, $X_{cls}$\} for training prior networks, encoder and decoder. Taking $X_{mt}$ for example, we follow~\citet{zhang-etal-2016-variational} and apply \emph{mean-pooling} over the output $\mathbf{h}^{N_e,X_{mt}}_{e}$ of the $N_e$-th encoder layer:
\begin{equation}
\setlength{\abovedisplayskip}{5pt}
\setlength{\belowdisplayskip}{5pt}
\begin{split}
\mathbf{h}_{X_{mt}}=\frac{1}{|X_{mt}|}\sum_{i=1}^{|X_{mt}|}(\mathbf{h}^{N_e,X_{mt}}_{e,i}). \nonumber
\end{split}
\end{equation}
Similarly, we obtain $\mathbf{h}_{X_{ms}^{src}}$ and $\mathbf{h}_{X_{cls}}$.

For training recognition networks, we obtain the representations of \{$Y_{mt}$, $X_{ms}^{tgt}$, $Y_{cls}$\}, taking $Y_{mt}$ for example, and calculate it as follows:
\begin{equation}
\setlength{\abovedisplayskip}{5pt}
\setlength{\belowdisplayskip}{5pt}
\begin{split}
\mathbf{h}_{Y_{mt}}=\frac{1}{|Y_{mt}|}\sum_{i=1}^{|Y_{mt}|}(\mathbf{h}^{N_e,Y_{mt}}_{e,i}). \nonumber
\end{split}
\end{equation}
Similarly, we obtain $\mathbf{h}_{X_{ms}^{tgt}}$ and $\mathbf{h}_{Y_{cls}}$. 

\textbf{\begin{figure}[t]
    \centering
    \includegraphics[width=0.49\textwidth]{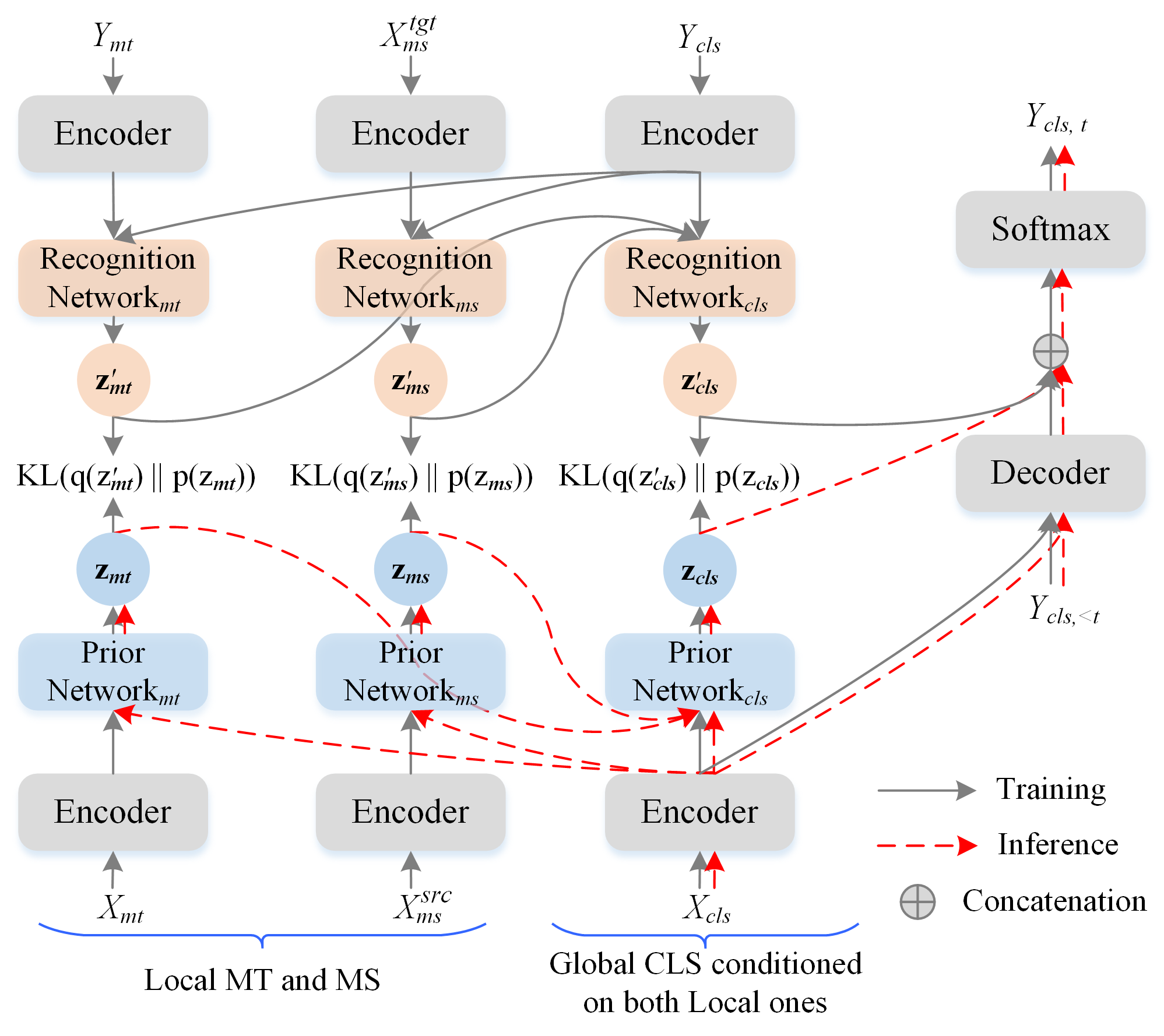}
    \caption{Overview of the proposed VHM framework. The local variables $\mathbf{z}_{mt}$, $\mathbf{z}_{ms}$ are tailored for translation and summarization, respectively. Then the global one $\mathbf{z}_{cls}$ is for cross-lingual summarization, where the $\mathbf{z}_{cls}$ not only conditions on the input but also $\mathbf{z}_{mt}$ and $\mathbf{z}_{ms}$. The solid grey lines indicate training process responsible for generating \{$\mathbf{z}_{mt}^{\prime},\mathbf{z}_{ms}^{\prime},\mathbf{z}_{cls}^{\prime}$\} from the corresponding posterior distribution predicted by recognition networks, which guide the learning of prior networks. The dashed red lines indicate inference process for generating \{$\mathbf{z}_{mt},\mathbf{z}_{ms},\mathbf{z}_{cls}$\} from the corresponding prior distributions predicted by prior networks. The encoder is shared by all tasks with a bilingual vocabulary.
    }
    \label{fig:model}
\end{figure}}

\subsection{Variational Hierarchical Modules}
Firstly, we design two local latent variational modules to learn the translation distribution in MT pairs and summarization distribution in MS pairs, respectively. Then, conditioned on them, we introduce a global latent variational module to explicitly exploit them.
\label{sec:rsc}
\subsubsection{Local: Translation and Summarization}
\paragraph{Translation.} To capture the translation of the paired sentences, we introduce a local variable $\mathbf{z}_{mt}$ that is responsible for generating the target information. Inspired by~\citet{ijcai2019-727}, we use isotropic Gaussian distribution as the prior distribution of $\mathbf{z}_{mt}$: $p_\theta(\mathbf{z}_{mt}|X_{mt}) \sim \mathcal{N}(\bm{\mu}_{mt}, \bm{\sigma}_{mt}^2\mathbf{I})$, where $\mathbf{I}$ denotes the identity matrix and we have
\begin{equation}
\setlength{\abovedisplayskip}{5pt}
\setlength{\belowdisplayskip}{5pt}
\begin{split}
        \bm{\mu}_{mt}&= \mathrm{MLP}_\theta^{mt}(\mathbf{h}_{X_{mt}}),  \\ 
        \bm{\sigma}_{mt} &=\mathrm{Softplus}(\mathrm{MLP}_\theta^{mt}(\mathbf{h}_{X_{mt}})), 
\end{split}
\end{equation}
where $\mathrm{MLP(\cdot)}$ and $\mathrm{Softplus(\cdot)}$ are multi-layer perceptron and approximation of
$\mathrm{ReLU}$ function, respectively. 

At training, the posterior distribution conditions on both source input and the target reference, which provides translation information. Therefore, the prior network can learn a tailored translation distribution by approaching the recognition network via $\mathrm{KL}$ divergence~\cite{kingma2013auto}: $q_\phi(\mathbf{z}_{mt}^{\prime}|X_{mt},Y_{mt})\sim \mathcal{N}({\bm \mu}_{mt}^\prime, {\bm \sigma}_{mt}^{{\prime}2}\mathbf{I})$, where ${\bm \mu}_{mt}^\prime$ and $\bm{\sigma}_{mt}^{\prime}$ are calculated as:
\begin{equation}
\setlength{\abovedisplayskip}{5pt}
\setlength{\belowdisplayskip}{5pt}
\begin{split}
        \bm{\mu}^\prime_{mt}&= \mathrm{MLP}_\phi^{mt}(\mathbf{h}_{X_{mt}}; \mathbf{h}_{Y_{mt}}), \\ 
        \bm{\sigma}^\prime_{mt} &=\mathrm{Softplus}(\mathrm{MLP}_\phi^{mt}(\mathbf{h}_{X_{mt}}; \mathbf{h}_{Y_{mt}})), 
\end{split}
\end{equation}
where ($\cdot$;$\cdot$) indicates concatenation operation.
\label{sec:lwc}
\paragraph{Summarization.}
To capture the summarization in MS pairs, we introduce another local variable $\mathbf{z}_{ms}$, which takes charge of generating the source-language summary. Similar to $\mathbf{z}_{mt}$, we define its prior distribution as: $p_\theta(\mathbf{z}_{ms}|X_{ms}^{src})\sim \mathcal{N}(\bm{\mu}_{ms}, \bm{\sigma}_{ms}^2\mathbf{I})$, where $\bm{\mu}_{ms}$ and $\bm{\sigma}_{ms}$ are calculated as:
\begin{equation}
\setlength{\abovedisplayskip}{5pt}
\setlength{\belowdisplayskip}{5pt}
\begin{split}
        \bm{\mu}_{ms}&= \mathrm{MLP}_\theta^{ms}(\mathbf{h}_{X_{ms}^{src}}),  \\ 
        \bm{\sigma}_{ms} &=\mathrm{Softplus}(\mathrm{MLP}_\theta^{ms}(\mathbf{h}_{X_{ms}^{src}})). 
\end{split}
\end{equation}

At training, the posterior distribution conditions on both the source input and the source-language summary that contains the summarization clue, and thus is responsible for guiding the learning of the prior distribution. Specifically, we define the posterior distribution as: $q_\phi(\mathbf{z}_{ms}^{\prime}|X_{ms}^{src}, X_{ms}^{tgt})\sim \mathcal{N}(\bm{\mu}_{ms}^\prime, \bm{\sigma}_{ms}^{\prime2}\mathbf{I})$, where $\bm{\mu}_{ms}^\prime$ and $\bm{\sigma}_{ms}^{\prime}$ are calculated as:
\begin{equation}
\setlength{\abovedisplayskip}{5pt}
\setlength{\belowdisplayskip}{5pt}
\begin{split}
        \bm{\mu}^\prime_{ms}&= \mathrm{MLP}_\phi^{ms}(\mathbf{h}_{X_{ms}^{src}}; \mathbf{h}_{X_{ms}^{tgt}}),  \\ 
        \bm{\sigma}^\prime_{ms} &=\mathrm{Softplus}(\mathrm{MLP}_\phi^{ms}(\mathbf{h}_{X_{ms}^{src}}; \mathbf{h}_{X_{ms}^{tgt}})). 
\end{split}
\end{equation}

\subsubsection{Global: CLS}
After obtaining $\mathbf{z}_{mt}$ and $\mathbf{z}_{ms}$, we introduce the global variable $\mathbf{z}_{cls}$ that aims to generate a target-language summary, where the $\mathbf{z}_{cls}$ can simultaneously exploit the local variables for CLS. Specifically, we firstly encode the source input $X_{cls}$ and condition on both two local variables $\mathbf{z}_{mt}$ and $\mathbf{z}_{ms}$, and then sample $\mathbf{z}_{cls}$. We define its prior distribution as: $p_\theta(\mathbf{z}_{cls}|X_{cls},\mathbf{z}_{mt},\mathbf{z}_{ms})\sim \mathcal{N}(\bm{\mu}_{cls}, \bm{\sigma}_{cls}^2\mathbf{I})$, where $\bm{\mu}_{cls}$ and $\bm{\sigma}_{cls}$ are calculated as:
\begin{equation}
\setlength{\abovedisplayskip}{5pt}
\setlength{\belowdisplayskip}{5pt}
\begin{split}
        \bm{\mu}_{cls}&= \mathrm{MLP}_\theta^{cls}(\mathbf{h}_{X_{cls}}; \mathbf{z}_{mt}; \mathbf{z}_{ms}), \\
        \bm{\sigma}_{cls} &=\mathrm{Softplus}(\mathrm{MLP}_\theta^{cls}(\mathbf{h}_{X_{cls}}; \mathbf{z}_{mt}; \mathbf{z}_{ms})). 
\end{split}
\end{equation}

At training, the posterior distribution conditions on the local variables, the CLS input, and the cross-lingual summary that contains combination information of translation and summarization. Therefore, the posterior distribution can teach the prior distribution. Specifically, we define the posterior distribution as: $q_\phi(\mathbf{z}_{cls}^{\prime}|X_{cls},\mathbf{z}_{mt},\mathbf{z}_{ms},Y_{cls})\sim \mathcal{N}(\bm{\mu}_{cls}^\prime, \bm{\sigma}_{cls}^{\prime2}\mathbf{I})$, where $\bm{\mu}_{cls}^\prime$ and $\bm{\sigma}_{cls}^{\prime}$ are calculated as:
\begin{equation}
\setlength{\abovedisplayskip}{5pt}
\setlength{\belowdisplayskip}{5pt}
\resizebox{.99\hsize}{!}{$
\begin{split}
        \bm{\mu}^\prime_{cls}&= \mathrm{MLP}_\phi^{cls}(\mathbf{h}_{X_{cls}}; \mathbf{z}_{mt}; \mathbf{z}_{ms}; \mathbf{h}_{Y_{cls}}), \\ 
        \bm{\sigma}^\prime_{cls} &=\mathrm{Softplus}(\mathrm{MLP}_\phi^{cls}(\mathbf{h}_{X_{cls}}; \mathbf{z}_{mt}; \mathbf{z}_{ms}; \mathbf{h}_{Y_{cls}})). 
\end{split}$}
\end{equation}

\subsection{Decoder}
\label{sec:dec}
The decoder adopts a similar structure to the encoder, and each of $N_d$ decoder layers includes an additional cross-attention sub-layer ($\mathrm{CrossAtt}$):
\begin{equation}
\setlength{\abovedisplayskip}{5pt}
\setlength{\belowdisplayskip}{5pt}
\begin{split}
\label{eq:trans_de}
    \mathbf{s}^\ell_d &= \mathrm{SelfAtt}(\mathbf{h}^{\ell-1}_d) + \mathbf{h}^{\ell-1}_d, \nonumber\\
    \mathbf{c}^\ell_d &= \mathrm{CrossAtt}(\mathbf{s}^{\ell}_d, \mathbf{h}_e^{N_e}) + \mathbf{s}^\ell_d, \\
    \mathbf{h}^\ell_d &= \mathrm{FFN}(\mathbf{c}^\ell_d) + \mathbf{c}^\ell_d, \nonumber
\end{split}
\end{equation}
where $\mathbf{h}^{\ell}_d$ denotes the state of the $\ell$-th decoder layer.

As shown in \autoref{fig:model}, we firstly obtain the local two variables either from the posterior distribution predicted by recognition networks (training process as the solid grey lines) or from prior distribution predicted by prior networks (inference process as the dashed red lines). Then, conditioned on the local two variables, we generate the global variable ($\mathbf{z}_{cls}^{\prime}$/$\mathbf{z}_{cls}$) via posterior (training) or prior (inference) network. Finally, we incorporate $\mathbf{z}_{cls}^{(\prime)}$\footnote{Here, we use $\mathbf{z}_{cls}^{\prime}$ when training and $\mathbf{z}_{cls}$ during inference, as similar to \autoref{eq:output}.} into the state of the top layer of the decoder with a projection layer:
\begin{equation}
\setlength{\abovedisplayskip}{5pt}
\setlength{\belowdisplayskip}{5pt}
\begin{split}
\label{eq:output}
    \mathbf{o}_t &= \mathrm{Tanh}(\mathbf{W}_p[\mathbf{h}^{N_d}_{d,t}; \mathbf{z}_{cls}^{(\prime)}] + \mathbf{b}_p), 
\end{split}
\end{equation}
where $\mathbf{W}_p$ and $\mathbf{b}_p$ are training parameters, $\mathbf{h}^{N_d}_{d,t}$ is the hidden state at time-step $t$ of the $N_d$-th decoder layer. Then, $\mathbf{o}_t$ is fed into a linear transformation and softmax layer to predict the probability distribution of the next target token:
\begin{equation}
\setlength{\abovedisplayskip}{5pt}
\setlength{\belowdisplayskip}{5pt}
\begin{split}
    \mathbf{p}_t &= \mathrm{Softmax}(\mathbf{W}_o\mathbf{o}_t+\mathbf{b}_o),\nonumber
\end{split}
\end{equation}
where $\mathbf{W}_o$ and $\mathbf{b}_o$ are training parameters.
\subsection{Training and Inference}
\label{sec:to}
The model is trained to maximize the conditional log-likelihood, due to the intractable marginal likelihood, which is converted to the following varitional lower bound that needs to be maximized in the training process:
\begin{equation}\nonumber
\setlength{\abovedisplayskip}{5pt}
\setlength{\belowdisplayskip}{5pt}
\resizebox{.99\hsize}{!}{$
\begin{split}
    &\mathcal{J}(\theta,\phi;X_{cls},X_{mt},X_{ms}^{src},Y_{cls},Y_{mt},X_{ms}^{tgt}) =\\
    &-\mathrm{KL}(q_\phi (\mathbf{z}_{mt}^{\prime}|X_{mt},Y_{mt}) \| p_\theta (\mathbf{z}_{mt}|X_{mt})) \\
    &-\mathrm{KL}(q_\phi (\mathbf{z}_{ms}^{\prime}|X_{ms}^{src},X_{ms}^{tgt}) \| p_\theta (\mathbf{z}_{ms}|X_{ms}^{src})) \\
    &-\mathrm{KL}(q_\phi (\mathbf{z}_{cls}^{\prime}|X_{cls},\mathbf{z}_{mt},\mathbf{z}_{ms},Y_{cls}) \| p_\theta (\mathbf{z}_{cls}|X_{cls},\mathbf{z}_{mt},\mathbf{z}_{ms})) \\
    &+\mathbb{E}_{q_\phi} [\mathrm{log} p_\theta(Y_{mt}|X_{mt},\mathbf{z}_{mt})]\\
    &+\mathbb{E}_{q_\phi} [\mathrm{log} p_\theta(X_{ms}^{tgt}|X_{ms}^{src},\mathbf{z}_{ms})]\\
    &+\mathbb{E}_{q_\phi} [\mathrm{log} p_\theta(Y_{cls}|X_{cls},\mathbf{z}_{cls},\mathbf{z}_{mt},\mathbf{z}_{ms})],
\end{split}
$}
\end{equation}
where the variational lower bound includes the reconstruction terms and KL divergence terms based on three hierarchical variables.
We use the reparameterization trick~\cite{kingma2013auto} to estimate the gradients of the prior and recognition networks~\cite{zhao-etal-2017-learning}.

During inference, firstly, the prior networks of MT and MS generate the local variables. Then, conditioned on them, the global variable is produced by prior network of CLS. Finally, only the global variable is fed into the decoder, which corresponds to red dashed arrows in~\autoref{fig:model}.

\section{Experiments}
\subsection{Datasets and Metrics}

\paragraph{Datasets.}
We evaluate the proposed approach on Zh2EnSum and En2ZhSum datasets released by~\cite{zhu-etal-2019-ncls}.\footnote{\url{https://github.com/ZNLP/NCLS-Corpora}} The Zh2EnSum and En2ZhSum are originally from~\cite{lcsts} and~\cite{hermann2015,Zhu2018MSMO}, respectively. Both the Chinese-to-English and English-to-Chinese test sets are manually corrected. The involved training data in our experiments are listed in~\autoref{tbl:involved1}.

\begin{table}[t!]
\centering
\newcommand{\tabincell}[2]{\begin{tabular}{@{}#1@{}}#2\end{tabular}}
\scalebox{0.80}{
\setlength{\tabcolsep}{0.9mm}{
\begin{tabular}{c|l|lll}
\hline
\multirow{3}{*}{\tabincell{c}{Zh2EnSum}}
&D1&CLS&{Zh2EnSum}   &1,693,713    \\ 
&D2&MS&{LCSTS}   &1,693,713   \\ 
&D3&MT&{LDC}  &2.08M   \\ \hline
\multirow{3}{*}{\tabincell{c}{En2ZhSum}}
&D4&CLS& {En2ZhSum}  &364,687  \\ 
&D5&MS& ENSUM   &364,687        \\
&D3&MT&{LDC}  &2.08M      \\
\hline
\end{tabular}}}
\caption{Involved training data. LCSTS~\cite{lcsts} is a Chinese summarization dataset. LDC corpora includes LDC2000T50, LDC2002L27, LDC2002T01,
LDC2002E18, LDC2003E07, LDC2003E14, LDC2003T17,
and LDC2004T07. ENSUM consists of {CNN/Dailymail~\cite{hermann2015} and MSMO~\cite{Zhu2018MSMO}}. }
\label{tbl:involved1}
\end{table}

\begin{table}[t!]
\small
\centering
\begin{tabular}{@{}l|ccc@{}}
\hline
\multirow{2}{*}{\textbf{Models}} & \multicolumn{3}{c}{\textbf{Zh2EnSum}} \\ \cline{2-4} 
 & Size (M) & Train (S) & Data\\ 
 \hline
ATS-A  & 137.60 & 30  & D1\&D3  \\
MS-CLS & 211.41 & 48 & D1\&D2  \\
MT-CLS & 208.84 & 63 & D1\&D3 \\
MT-MS-CLS& 114.90 & 24 & D1\&D2\&D3 \\
VHM & 117.40 & 27 & D1\&D2\&D3 \\ 
\hline
\end{tabular}
\caption{Model details. Size (M): number of trainable parameters; Train (S) denotes how many seconds required for each model to train the 100-batch cross-lingual summarization task of the same batch size (3072). Data: Training Data, as listed in~\autoref{tbl:involved1}.}
\label{tbl:involved2}
\end{table}

\begin{table}[t!]
\small
\centering
\begin{tabular}{@{}l|ccc@{}}
\hline
\multirow{2}{*}{\textbf{Models}} & \multicolumn{3}{c}{\textbf{En2ZhSum}} \\ \cline{2-4} 
 & Size (M) & Train (S) & Data\\ 
 \hline
ATS-A  & 115.05 & 25  & D4\&D3  \\
MS-CLS & 190.23 & 65 & D4\&D5  \\
MT-CLS & 148.16 & 72 & D4\&D3 \\
MT-MS-CLS& 155.50 & 32 & D4\&D5\&D3 \\
VHM & 158.00 & 36 & D4\&D5\&D3 \\ 
\hline
\end{tabular}
\caption{Model details. Size (M): number of trainable parameters; Train (S) denotes how many seconds required for each model to train the 100-batch cross-lingual summarization task of the same batch size (3072). Data: Training Data, as listed in~\autoref{tbl:involved1}.}
\label{tbl:involved3}
\end{table}

\noindent \textbf{Zh2EnSum.}
It is a Chinese-to-English summarization dataset, which has 1,699,713 Chinese short texts (104 Chinese characters on average) paired with Chinese (18 Chinese characters on average) and English short summaries (14 tokens on average). The dataset is split into 1,693,713 training pairs, 3,000 validation pairs, and 3,000 test pairs. The involved training data used in multi-task learning, model size, training time, are listed in~\autoref{tbl:involved2}.

\noindent \textbf{En2ZhSum.} 
It is an English-to-Chinese summarization dataset, which has 370,687 English documents (755 tokens on average) paired with multi-sentence English (55 tokens on average) and Chinese summaries (96 Chinese characters on average). The dataset is split into 364,687 training pairs, 3,000 validation pairs, and 3,000 test pairs. The involved training data used in multi-task learning, model size, training time, are listed in~\autoref{tbl:involved3}.

\begin{table*}[t!]
\centering
\newcommand{\tabincell}[2]{\begin{tabular}{@{}#1@{}}#2\end{tabular}}
\setlength{\tabcolsep}{1.1mm}{
\begin{tabular}{l|l|llll|lll}
\hline
\multirow{2}{*}{M\#}&\multicolumn{1}{c|}{\multirow{2}{*}{\textbf{Models}}} &\multicolumn{4}{c|}{\textbf{Zh2EnSum}}    &\multicolumn{3}{c}{\textbf{En2ZhSum}} \\ 
\cline{3-6} \cline{7-9}
&\multicolumn{1}{l|}{} & \multicolumn{1}{l}{RG1} & \multicolumn{1}{l}{RG2} & \multicolumn{1}{l}{RGL} &  \multicolumn{1}{l|}{MVS}      & \multicolumn{1}{l}{RG1} & \multicolumn{1}{l}{RG2} & \multicolumn{1}{l}{RGL}  \\ \hline
M1&{GETran}~\cite{zhu-etal-2019-ncls}    & 24.34 &9.14 &20.13 &0.64      & 28.19 &11.40 &25.77    \\
M2&{GLTran}~\cite{zhu-etal-2019-ncls}   &35.45 &16.86 &31.28 &16.90 &32.17 &13.85 &29.43\\\cdashline{1-9}[4pt/2pt]
M3&{TNCLS}~\cite{zhu-etal-2019-ncls}    &38.85 &21.93 &35.05 &19.43     &36.82 &18.72 &33.20  \\
M4&{ATS-A}~\cite{zhu-etal-2020-attend}   &40.68 &24.12 &36.97 &22.15 &40.47 &22.21 &36.89 \\\cdashline{1-9}[4pt/2pt]
M5&{MS-CLS}~\cite{zhu-etal-2019-ncls}  &40.34 &22.65 &36.39 &21.09   &38.25 &20.20 &34.76  \\
M6&{MT-CLS}~\cite{zhu-etal-2019-ncls}  & 40.25 &22.58 &36.21 &21.06 & 40.23 &22.32 &36.59   \\ 
M7&{MS-CLS-Rec}~\cite{cao-etal-2020-jointly}&40.97&23.20&36.96 &NA &38.12& 16.76 &33.86\\
M8&{MS-CLS}*  &40.44	&22.19	&36.32	&21.01   &38.26     &20.07 & 34.49  \\
M9&{MT-CLS}*  &40.05	&21.72	&35.74	&20.96  &40.14      &22.36  & 36.45   \\
M10&{MT-MS-CLS} (Ours)  &40.65	&24.02	&36.69	&22.17  &{40.34}     &{22.35} &36.44  \\
M11&{VHM} (Ours)   &{41.36}$^{\dagger\dagger}$  &{24.64}$^{\dagger}$ &{37.15}$^{\dagger}$  &{22.55}$^{\dagger}$ &{40.98}$^{\dagger\dagger}$  &{23.07}$^{\dagger\dagger}$  &{37.12}$^{\dagger}$\\ \cdashline{1-9}[4pt/2pt] 
M12&mBART~\cite{liu-etal-2020-multilingual-denoising}      &43.61	&25.14	&38.79	&23.47 &41.55 &23.27  &37.22  \\
M13&MLPT~\cite{xu-etal-2020-mixed} &{43.50} &25.41 &29.66 &NA	&41.62 &23.35 &37.26\\
M14&{VHM + mBART} (Ours)  &\textbf{43.97}$^{\dagger}$  &\textbf{25.61}$^{\dagger}$ &\textbf{39.19}$^{\dagger}$  &\textbf{23.88} &\textbf{41.95}$^{\dagger}$  &\textbf{23.54}$^{\dagger}$  &\textbf{37.67}$^{\dagger}$\\ 
\hline
\end{tabular}}
\caption{ROUGE F1 scores (\%) and MoverScore scores (\%) on Zh2EnSum test set, and ROUGE F1 scores (\%) on En2ZhSum test set. RG and MVS refer to ROUGE and MoverScore, respectively. The ``*'' denotes results by running their released code. The ``NA'' indicates no such result in the original paper. ``$^{\dagger}$'' and ``$^{\dagger\dagger}$'' indicate that statistically significant better (M11 vs. M4 and M14 vs. M12) with t-test {\em p} \textless \ 0.05 and {\em p} \textless \ 0.01, respectively. ``VHM + mBART'' means that we use mBART weights as model initialization of our VHM.}
\label{tbl:main_res}
\end{table*}

\paragraph{Metrics.}

Following~\citet{zhu-etal-2020-attend}, 1) we evaluate all models with the standard ROUGE metric~\cite{lin-2004-rouge}, reporting the F1 scores for ROUGE-1, ROUGE-2, and ROUGE-L. All ROUGE scores are reported by the 95\% confidence interval measured by the official script;\footnote{The parameter for ROUGE script here is ``-c 95 -r 1000 -n 2 -a''} 2) we also evaluate the quality of English summaries in Zh2EnSum with MoverScore~\cite{zhao-etal-2019-moverscore}.

\subsection{Implementation Details}
In this paper, we train all models using standard transformer~\cite{vaswani2017attention} in \emph{Base} setting. For other hyper-parameters, we mainly follow the setting described in~\citet{zhu-etal-2019-ncls,zhu-etal-2020-attend} for fair comparison. For more details, please refer to~\autoref{ID}.

\subsection{Comparison Models}
\paragraph{Pipeline Models.}

 {TETran}~\cite{zhu-etal-2019-ncls}. It first translates the original article into the target language by Google Translator\footnote{\url{https://translate.google.com/}} and then summarizes the translated text via LexRank~\cite{erkan2004lexrank}. {TLTran}~\cite{zhu-etal-2019-ncls}. It first summarizes the original article via a transformer-based monolingual summarization model and then translates the summary into the target language by Google Translator.

\paragraph{End-to-End Models.}

 {TNCLS}~\cite{zhu-etal-2019-ncls}. It directly uses the de-facto transformer~\cite{vaswani2017attention} to train an end-to-end CLS system. {ATS-A}~\cite{zhu-etal-2020-attend}.\footnote{\url{https://github.com/ZNLP/ATSum}} It is an efficient model to attend the pre-constructed probabilistic bilingual lexicon to enhance the CLS. 
 {MS-CLS}~\cite{zhu-etal-2019-ncls}. It simultaneously performs summarization generation for both CLS and MS tasks and calculates the total losses. {MT-CLS}~\cite{zhu-etal-2019-ncls}.\footnote{\url{https://github.com/ZNLP/NCLS-Corpora}} It alternatively trains CLS and MT tasks. {MS-CLS-Rec}~\cite{cao-etal-2020-jointly}. It jointly trains MS and CLS systems with a reconstruction loss to mutually map the source and target representations. mBART~\cite{liu-etal-2020-multilingual-denoising}. We use mBART ($mbart.cc25$) as model initialization to fine-tune the CLS task. MLPT (Mixed-Lingual Pre-training)~\cite{xu-etal-2020-mixed}. It applies mixed-lingual pretraining that leverages six related tasks, covering both cross-lingual tasks such as translation and monolingual tasks like masked language models. {MT-MS-CLS}. It is our strong baseline, which is implemented by alternatively training CLS, MT, and MS. Here, we keep the dataset used for MT and MS consistent with~\citet{zhu-etal-2019-ncls} for fair comparison.

\begin{figure*}[ht]
    \centering
    \includegraphics[width=0.99\textwidth]{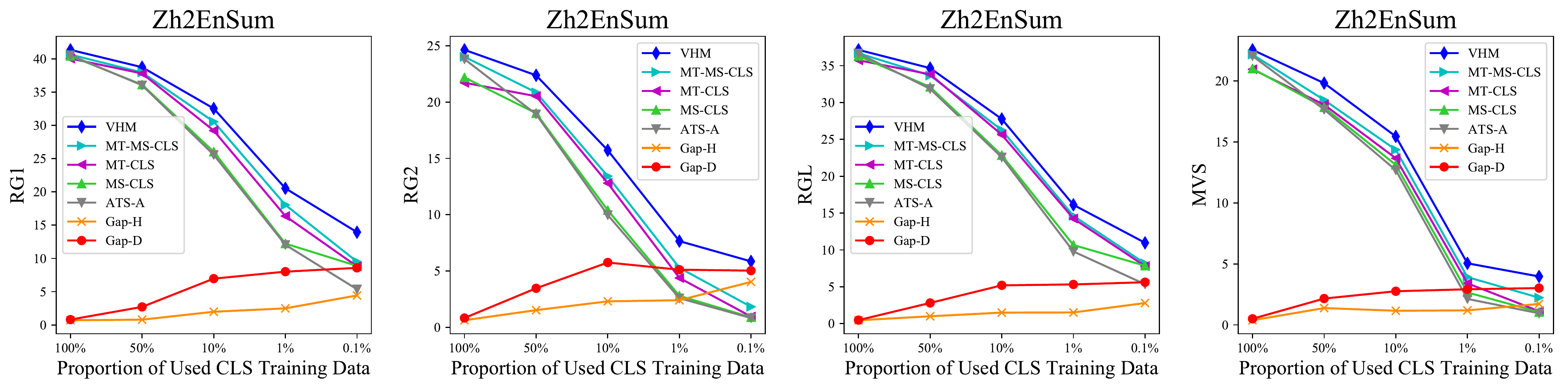}
    \caption{ROUGE F1 scores (\%) and MoverScore scores (\%) on Zh2EnSum test set in few-shot setting. \textbf{x\%} means that the \textbf{x\%} CLS training dataset is used, \emph{e.g.}, \textbf{0.1\%} represents that \textbf{0.1\%} training dataset (about 1.7k instances) is used for training. The performance ``Gap-H'' (orange line) between ``VHM'' and ``MT-MS-CLS'' grows steadily with the decreasing of used CLS training data, which is similar to the performance ``Gap-D'' (red line) between ``VHM'' and ``ATS-A''. 
    }
    \label{fig:gap1}
\end{figure*}

\subsection{Main Results}
Overall, we separate the models into three parts in \autoref{tbl:main_res}: the pipeline, end-to-end, and multi-task settings. In each part, we show the results of existing studies and our re-implemented baselines and our approach, \emph{i.e.}, the VHM, on Zh2EnSum and En2ZhSum test sets. 
\begin{figure*}[ht]
    \centering
    \includegraphics[width=0.74\textwidth]{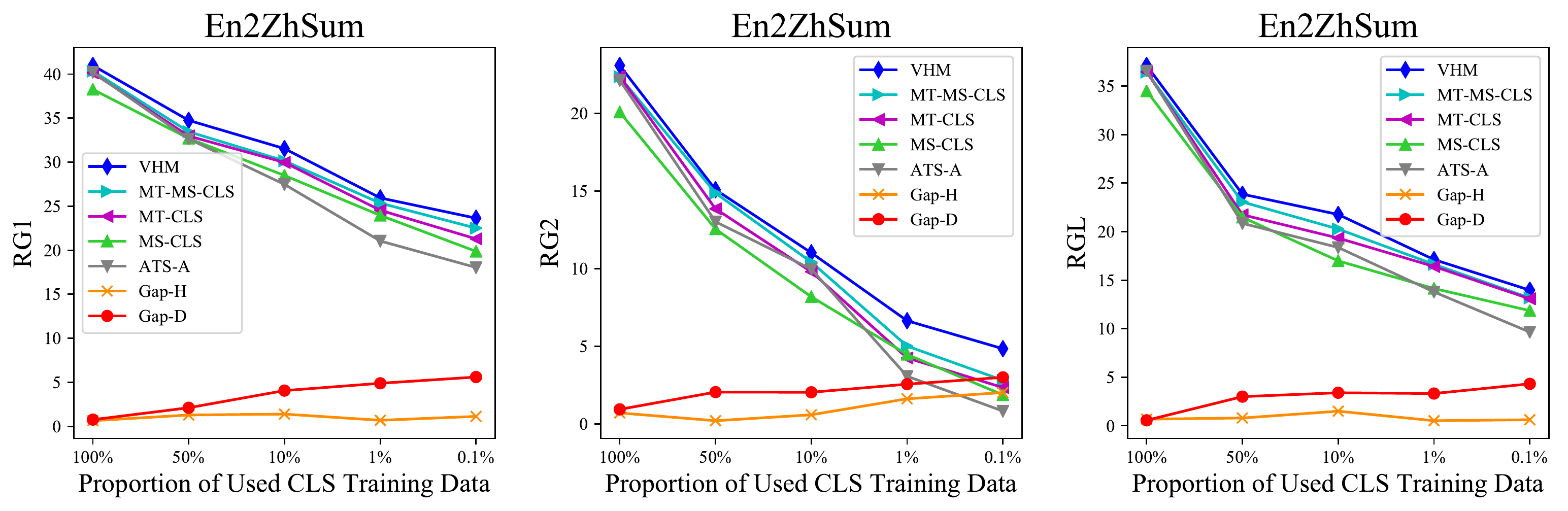}
    \caption{Rouge F1 scores (\%) on the test set when using different CLS training data. The performance ``Gap-H'' (orange line) between ``VHM'' and ``MT-MS-CLS'' grows steadily with the decreasing of used CLS training data on ROUGE-2, which is similar to the performance ``Gap-D'' (red line) between ``VHM'' and ``ATS-A''. 
    }
    \label{fig:gap2}
\end{figure*}
\paragraph{Results on Zh2EnSum.}
Compared against the pipeline and end-to-end methods, VHM substantially outperforms all of them (\emph{e.g.}, the previous best model ``ATS-A'') by a large margin with 0.68/0.52/0.18/0.4$\uparrow$ scores on RG1/RG2/RGL/MVS, respectively. Under the multi-task setting, compared to the existing best model ``MS-CLS-Rec'', our VHM also consistently boosts the performance in three metrics (\emph{i.e.}, 0.39$\uparrow$, 1.44$\uparrow$, and 0.19$\uparrow$ ROUGE scores on RG1/RG2/RGL, respectively), showing its effectiveness. Our VHM also significantly surpasses our strong baseline ``MT-MS-CLS'' by 0.71/0.62/0.46/0.38$\uparrow$ scores on RG1/RG2/RGL/MVS, respectively, demonstrating the superiority of our model again. 

After using mBART as model initialization, our VHM achieves the state-of-the-art results on all metrics.

\paragraph{Results on En2ZhSum.}
Compared against the pipeline, end-to-end and multi-task methods, our VHM presents remarkable ROUGE improvements over the existing best model ``ATS-A'' by a large margin, about 0.51/0.86/0.23$\uparrow$ ROUGE gains on RG1/RG2/RGL, respectively. These results suggest that VHM consistently performs well in different language directions.

Our approach still notably surpasses our strong baseline ``MT-MS-CLS'' in terms of all metrics, which shows the generalizability and superiority of our model again.

\subsection{Few-Shot Results}
Due to the difficulty of acquiring the cross-lingual summarization dataset~\cite{zhu-etal-2019-ncls}, we conduct such experiments to investigate the model performance when the CLS training dataset is limited, \emph{i.e.}, few-shot experiments. Specifically, we randomly choose 0.1\%, 1\%, 10\%, and 50\% CLS training datasets to conduct experiments. The results are shown in~\autoref{fig:gap1} and~\autoref{fig:gap2}. 

\paragraph{Results on Zh2EnSum.}~\autoref{fig:gap1} shows that VHM significantly surpasses all comparison models under each setting. Particularly, under the 0.1\% setting, our model still achieves best performances than all baselines, suggesting that our variational hierarchical model works well in the few-shot setting as well. Besides, we find that the performance gap between comparison models and VHM is growing when the used CLS training data become fewer. It is because relatively larger proportion of translation and summarization data are used, the influence from MT and MS becomes greater, effectively strengthening the CLS model.  Particularly, the performance “Gap-H” between MT-MS-CLS and VHM is also growing, where both models utilize the same data. This shows that the hierarchical relationship between MT\&MS and CLS makes substantial contributions to the VHM model in terms of four metrics. Consequently, our VHM achieves a comparably stable performance.

\paragraph{Results on En2ZhSum.}From~\autoref{fig:gap2}, we observe the similar findings on Zh2EnSum. This shows that VHM significantly outperforms all comparison models under each setting, showing the generalizability and superiority of our model again in the few-shot setting. 

\begin{table}[t!]
\centering
\scalebox{0.83}{
\setlength{\tabcolsep}{0.30mm}{
\begin{tabular}{l|l|c|ccc}
\hline
\multirow{2}{*}{\#}&\multicolumn{1}{c|}{\multirow{2}{*}{\textbf{Models}}} &\multicolumn{1}{c|}{\textbf{Zh2EnSum}}  &  \multicolumn{1}{c}{\textbf{En2ZhSum}}    \\ 
\cline{3-4}
&\multicolumn{1}{c|}{} & \multicolumn{1}{c|}{RG1/RG2/RGL/MVS} & \multicolumn{1}{c}{RG1/RG2/RGL}   \\ 
\hline
0&VHM & 41.36/24.64/37.15/22.55 & 40.98/23.07/37.12 \\\cdashline{1-6}[4pt/2pt]
1& -- $\mathbf{z}_{mt}$ & 40.75/23.47/36.48/22.18 & 40.35/22.48/36.55\\
2&-- $\mathbf{z}_{ms}$ & 40.69/23.34/36.35/22.12 & 40.57/22.79/36.71\\
3&-- $\mathbf{z}_{mt}$\&$\mathbf{z}_{ms}$ & 40.45/22.97/36.03/22.36 & 39.98/21.91/36.33\\
4&-- $\mathbf{z}_{cls}$ & 39.77/22.41/34.87/21.62 & 39.76/21.69/35.99\\
5&-- hierarchy & 40.47/22.64/34.96/21.78 & 39.67/21.79/35.87\\
\hline
\end{tabular}}}
\caption{Ablation results (in the full setting). Row 1 denotes that we remove the local variable $\mathbf{z}_{mt}$, and sample $\mathbf{z}_{cls}$ from the source input and another local variable $\mathbf{z}_{ms}$, similarly for row 2. Row 3 denotes that we remove both local variables $\mathbf{z}_{mt}$ and $\mathbf{z}_{ms}$ and sample $\mathbf{z}_{cls}$ only from the source input. Row 4 means that we remove the global variable $\mathbf{z}_{cls}$ and directly attend the local variables $\mathbf{z}_{mt}$ and $\mathbf{z}_{ms}$ in~\autoref{eq:output}. Row 5 represents that we keep three latent variables but remove the hierarchical relation between $\mathbf{z}_{cls}$ and $\mathbf{z}_{mt}$\&$\mathbf{z}_{ms}$.}
\label{abl}
\end{table}

\section{Analysis}
\subsection{Ablation Study}
\label{ssec:abs}
We conduct ablation studies to investigate how well the local and global variables of our VHM works. When removing variables listed in \autoref{abl}, we have the following findings. 

(1) Rows 1$\sim$3 vs. row 0 shows that the model performs worse, especially when removing the two local ones (row 3), due to missing the explicit translation or summarization or both information provided by the local variables, which is important to CLS. Besides, row 3 indicates that directly attending to $\mathbf{z}_{cls}$ leads to poor performances, showing the necessity of the hierarchical structure, \emph{i.e.}, using the global variable to exploit the local ones.

(2) Rows 4$\sim$5 vs. row 0 shows that directly attending the local translation and summarization cannot achieve good results due to lacking of the global combination of them, showing that it is very necessary for designing the variational hierarchical model, \emph{i.e.}, using a global variable to well exploit and combine the local ones.

\subsection{Human Evaluation} Following~\citet{zhu-etal-2019-ncls,zhu-etal-2020-attend}, we conduct human evaluation on 25 random samples from each of the Zh2EnSum and En2ZhSum test set. We compare the summaries generated by our methods (MT-MS-CLS and VHM) with the summaries generated by ATS-A, MS-CLS, and MT-CLS in the full setting and few-shot setting (0.1\%), respectively. We invite three graduate students to compare the generated summaries with human-corrected references, and assess each summary from three independent perspectives: 
\begin{enumerate}[itemindent=1em]
\item How \textbf{informative} (\emph{i.e.}, IF) the summary is? 
\item How \textbf{concise} (\emph{i.e.}, CC) the summary is? 
\item How \textbf{fluent}, grammatical (\emph{i.e.}, FL) the summary is? 
\end{enumerate}
Each property is assessed with a score from 1 (worst) to 5 (best). The average results are presented in~\autoref{human1} and ~\autoref{human2}. 

~\autoref{human1} shows the results in the full setting. We find that our VHM outperforms all comparison models from three aspects in both language directions, which further demonstrates the effectiveness and superiority of our model.

~\autoref{human2} shows the results in the few-shot setting, where only 0.1\% CLS training data are used in all models. We find that our VHM still performs best than all other models from three perspectives in both datasets, suggesting its generalizability and effectiveness again under different settings.

\begin{table}[]
\small
\centering
\begin{tabular}{@{}l|ccc|ccc@{}}
\hline
\multirow{2}{*}{\textbf{Models}} & \multicolumn{3}{c|}{\textbf{Zh2EnSum}} & \multicolumn{3}{c}{\textbf{En2ZhSum}} \\ \cline{2-7} 
 & IF & CC & FL & IF & CC & FL \\ 
\hline
ATS-A & 3.44 & 4.16 & 3.98 & 3.12 & 3.31 & 3.28 \\
MS-CLS & 3.12 & 4.08 & 3.76 & 3.04 & 3.22 & 3.12 \\
MT-CLS & 3.36 & 4.24 & 4.14 & 3.18 & 3.46 & 3.36 \\
MT-MS-CLS& 3.42 & 4.46 & 4.22 & 3.24 & 3.48 & 3.42 \\
VHM & \textbf{3.56} & \textbf{4.54} & \textbf{4.38} & \textbf{3.36} & \textbf{3.54} & \textbf{3.48} \\ 
\hline
\end{tabular}
\caption{Human evaluation results in the full setting. IF, CC and FL denote \textbf{informative}, \textbf{concise}, and \textbf{fluent} respectively.}
\label{human1}\vspace{-5pt}
\end{table}

\begin{table}[]
\small
\centering
\begin{tabular}{@{}l|ccc|ccc@{}}
\hline
\multirow{2}{*}{\textbf{Models}} & \multicolumn{3}{c|}{\textbf{Zh2EnSum}} & \multicolumn{3}{c}{\textbf{En2ZhSum}} \\ \cline{2-7} 
 & IF & CC & FL & IF & CC & FL \\ 
 \hline
ATS-A & 2.26 & 2.96 & 2.82 & 2.04 & 2.58 & 2.68 \\
MS-CLS & 2.24 & 2.84 & 2.78 & 2.02 & 2.52 & 2.64 \\
MT-CLS & 2.38 & 3.02 & 2.88 & 2.18 & 2.74 & 2.76 \\
MT-MS-CLS& 2.54 & 3.08 & 2.92 & 2.24 & 2.88 & 2.82 \\
VHM & \textbf{2.68} & \textbf{3.16} & \textbf{3.08} & \textbf{2.56} & \textbf{3.06} & \textbf{2.88} \\ 
\hline
\end{tabular}
\caption{Human evaluation results in the few-shot setting (0.1\%).}
\label{human2}\vspace{-10pt}
\end{table}

\section{Related Work}
\paragraph{Cross-Lingual Summarization.} Conventional cross-lingual summarization methods mainly focus on incorporating bilingual information into the pipeline methods~\cite{10.1145/979872.979877, ouyang-etal-2019-robust,orasan-chiorean-2008-evaluation,wan-etal-2010-cross,wan2011using,yao2015phrase,7502066}, \emph{i.e.}, translation and then summarization or summarization and then translation. Due to the difficulty of acquiring cross-lingual summarization dataset, some previous researches focus on constructing datasets~\cite{ladhak-etal-2020-wikilingua,scialom-etal-2020-mlsum,yela-bello-etal-2021-multihumes,zhu-etal-2019-ncls,DBLP:journals/corr/abs-2112-08804,perez-beltrachini-lapata-2021-models,varab-schluter-2021-massivesumm}, mixed-lingual pre-training~\cite{xu-etal-2020-mixed}, knowledge distillation~\cite{DBLP:journals/corr/abs-2112-03473}, contrastive  learning~\cite{wang-etal-2021-contrastive} or zero-shot approaches~\cite{liu2018zero,duan-etal-2019-zero,dou-etal-2020-deep}, \emph{i.e.}, using machine translation (MT) or monolingual summarization (MS) or both to train the CLS system. Among them,~\citet{zhu-etal-2019-ncls} propose to use roundtrip translation strategy to obtain large-scale CLS datasets and then present two multi-task learning methods for CLS. Based on this dataset,~\citet{zhu-etal-2020-attend} leverage an end-to-end model to attend the pre-constructed probabilistic bilingual lexicon to improve CLS. To further enhance CLS, some studies resort to shared decoder~\cite{bai-etal-2021-cross}, more pseudo training data~\cite{takase2020multitask}, or more related task training~\cite{Cao_Wan_Yao_Yu_2020,cao-etal-2020-jointly,DBLP:journals/corr/abs-2110-07936}. \citet{Wang2022ClidSumAB} concentrate on building a benchmark dataset for CLS on dialogue field. Different from them, we propose a variational hierarchical model that introduces a global variable to simultaneously exploit and combine the local translation variable in MT pairs and local summarization variable in MS pais for CLS, achieving better results.

\noindent \textbf{Conditional Variational Auto-Encoder.}\quad CVAE has verified its superiority in many fields~\cite{NIPS2015_8d55a249,liang-etal-2021-modeling,zhang-etal-2016-variational,SU2018287}. For instance, in dialogue, ~\citet{shen-etal-2019-modeling},~\citet{park-etal-2018-hierarchical} and \citet{AAAI1714567} extend CVAE to capture the semantic connection between the utterance and the corresponding context with hierarchical latent variables. Although the CVAE has been widely used in NLP tasks, its adaption and utilization to cross-lingual summarization for modeling hierarchical relationship are non-trivial, and to the best of our knowledge, has never been investigated before in CLS.

\noindent \textbf{Multi-Task Learning.} Conventional multi-task learning (MTL)~\cite{Multitask}, which trains the model on multiple related tasks to promote the representation learning and generalization performance, has been successfully used in NLP fields~\cite{10.1145/1390156.1390177,Deng2013NewTO,liang-etal-2021-towards,liang-etal-2021-iterative-multi,liang2020dependency}. In the CLS, conventional MTL has been explored to incorporate additional training data (MS, MT) into models~\cite{zhu-etal-2019-ncls,takase2020multitask,cao-etal-2020-jointly}. In this work, we instead focus on how to connect the relation between the auxiliary tasks at training to make the most of them for better CLS. 

\section{Conclusion}
In this paper, we propose to enhance the CLS model by simultaneously exploiting MT and MS. Given the hierarchical relationship between MT\&MS and CLS, we propose a variational hierarchical model to explicitly exploit and combine them in CLS process. Experiments on Zh2EnSum and En2ZhSum show that our model significantly improves the quality of cross-lingual summaries in terms of automatic metrics and human evaluations. Particularly, our model in the few-shot setting still works better, suggesting its superiority and generalizability. 

\section*{Acknowledgements}
The research work descried in this paper has been supported by the National Key R\&D Program of China (2020AAA0108001) and the National Nature Science Foundation of China (No. 61976015, 61976016, 61876198 and  61370130). Liang is supported by 2021 Tencent Rhino-Bird Research Elite Training Program. The authors would like to thank the anonymous reviewers for their valuable comments and suggestions to improve this paper.

\bibliography{anthology,custom}
\bibliographystyle{acl_natbib}

\appendix
\section*{Appendix}
\label{sec:appendix}
\section{Implementation Details}
\label{ID}
In this paper, we train all models using standard transformer~\cite{vaswani2017attention} in \emph{Base} setting. For other hyper-parameters, we mainly follow the setting described in~\cite{zhu-etal-2019-ncls,zhu-etal-2020-attend} for fair comparison. Specifically, the segmentation granularity is ``subword to subword'' for Zh2EnSum, and ``word to word'' for En2ZhSum. All the parameters are initialized via Xavier initialization method~\cite{Glorot10understandingthe}. We train our models using standard transformer~\cite{vaswani2017attention} in \emph{Base} setting, which contains a 6-layer encoder (\emph{i.e.}, $N_e$) and a 6-layer decoder (\emph{i.e.}, $N_d$) with 512-dimensional hidden representations. And all latent variables have a dimension of 128. Each mini-batch contains a set of document-summary pairs with roughly 4,096 source and 4,096 target tokens. We apply Adam optimizer~\cite{kingma2014adam} with $\beta_1$ = 0.9, $\beta_2$ = 0.998. Following~\citet{zhu-etal-2019-ncls}, we train each task for about 800,000 iterations in all multi-task models (reaching convergence). To alleviate the degeneration problem of the variational framework, we apply KL annealing. The KL multiplier $\lambda$ gradually increases from 0 to 1 over 400, 000 steps. All our methods without mBART as model initialization are trained and tested on a single NVIDIA Tesla V100 GPU. We use 8 NVIDIA Tesla V100 GPU to train our models when using mBART as model initialization, where the number of token on each GPU is set to 2,048 and the training step is set to 400, 000.

During inference, we use beam search with a beam size 4 and length penalty 0.6. 
\end{document}